\documentclass[10pt,twocolumn,letterpaper]{article}

\usepackage{cvpr}
\usepackage{times}
\usepackage{epsfig}
\usepackage{graphicx}
\usepackage{amsmath}
\usepackage{amssymb}
\usepackage{caption}
\usepackage{multirow}

\usepackage[pagebackref=true,breaklinks=true,letterpaper=true,colorlinks,bookmarks=false]{hyperref}

\cvprfinalcopy % *** Uncomment this line for the final submission

\def\cvprPaperID{4586} % *** Enter the CVPR Paper ID here

\newcommand{\dataset}{NoW\xspace}

\begin{document}

%%%%%%%%% TITLE
\title{Learning to Regress 3D Face Shape and Expression\\ 
from an Image without 3D Supervision}

%RingNet: Full Head Reconstruction from a Monocular Face Image with Robust and Accurate Geometry using Shape Consistency without 3D Supervision}

\author{Soubhik Sanyal \qquad Timo Bolkart \qquad Haiwen Feng \qquad Michael J. Black\\
Perceiving Systems Department\\
Max Planck Institute for Intelligent Systems\\
{\tt\small \{soubhik.sanyal, timo.bolkart, haiwen.feng, black\}@tuebingen.mpg.de}
% \and
}

\twocolumn[{
\renewcommand\twocolumn[1][]{#1}%
\maketitle
\begin{center}
	\newcommand{\teaserwidth}{\textwidth}
	\newcommand{\teaserheight}{0.666in}
	% \vspace{-0.35in}
	\centerline{
		\includegraphics[height=\teaserheight]{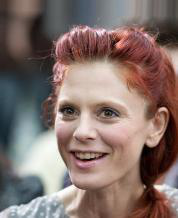} 
		\includegraphics[height=\teaserheight]{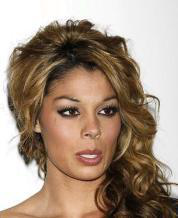} 
		\includegraphics[height=\teaserheight]{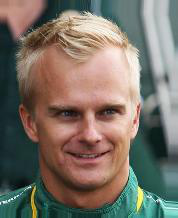} 
		\includegraphics[height=\teaserheight]{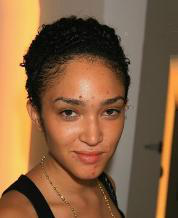} 
		\includegraphics[height=\teaserheight]{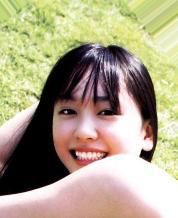} 
		\includegraphics[height=\teaserheight]{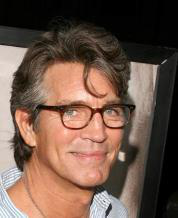} 
		\includegraphics[height=\teaserheight]{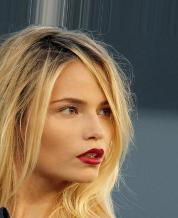} 
		\includegraphics[height=\teaserheight]{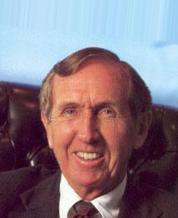} 
		\includegraphics[height=\teaserheight]{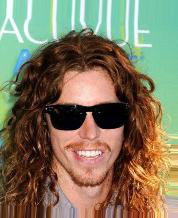} 
		\includegraphics[height=\teaserheight]{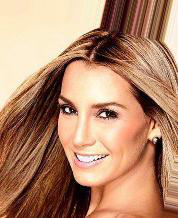} 
		\includegraphics[height=\teaserheight]{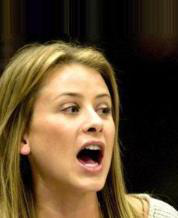} 
	}
	\centerline{
		\includegraphics[height=\teaserheight]{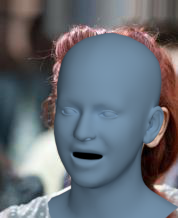} 
		\includegraphics[height=\teaserheight]{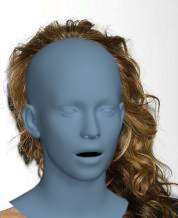} 
		\includegraphics[height=\teaserheight]{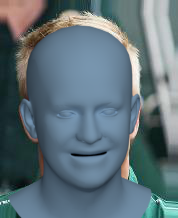} 
		\includegraphics[height=\teaserheight]{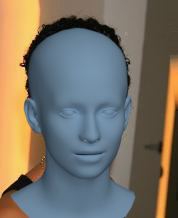} 
		\includegraphics[height=\teaserheight]{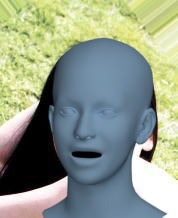} 
		\includegraphics[height=\teaserheight]{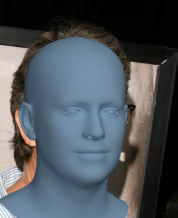} 
		\includegraphics[height=\teaserheight]{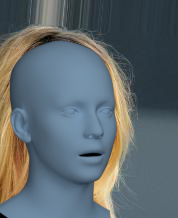} 
		\includegraphics[height=\teaserheight]{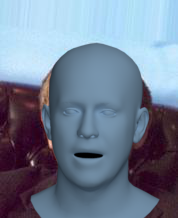} 
		\includegraphics[height=\teaserheight]{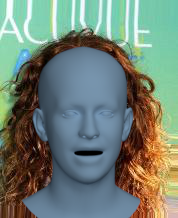} 
		\includegraphics[height=\teaserheight]{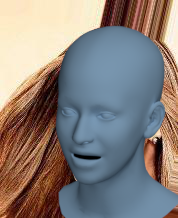} 
		\includegraphics[height=\teaserheight]{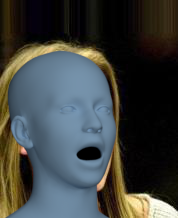} 
	}
	\captionof{figure}{{Without 3D supervision, \bf RingNet} learns a mapping from the
		pixels of a single image to the 3D facial parameters of the FLAME
		model \cite{Bolkart1}. Top: Images are from the CelebA dataset \cite{liu2015faceattributes}.  Bottom: estimated shape, pose and expression.
	}
	\vspace{-0.1in}
	\label{fig:teaser}
\end{center}%

}]

\maketitle
%\thispagestyle{empty}
%SMPL model notation

%shape
\newcommand{\shapecoeff}{\vec{\beta}}
\newcommand{\shapedim}{\vec{ \left| \beta \right|} }
\newcommand{\shapespace}{\mathcal{S}}
%pose
\newcommand{\posecoeff}{\vec{\theta}}
\newcommand{\posedim}{\vec{ \left| \theta \right|} }
\newcommand{\posespace}{\mathcal{P}}
%expression
\newcommand{\expcoeff}{\vec{\psi}}
\newcommand{\expdim}{\vec{ \left| \psi \right|}}
\newcommand{\expspace}{\mathcal{E}}
%template
\newcommand{\template}{\overline{\textbf{T}}}
%\newcommand{\template}{\textbf{T}}
%joints
\newcommand{\joints}{\textbf{J}}
\newcommand{\jointregressor}{\mathcal{J}}
%blendskinning
\newcommand{\blendweights}{\mathcal{W}}
\newcommand{\blendweightsdim}{\left| \mathcal{W} \right|}
%RingNet
\newcommand{\elemntsofring}{\textit{e}}
\newcommand{\numboferelements}{R}
\newcommand{\encoderoutput}{f_{\mathrm{enc}}}
\newcommand{\encoderoutputi}{f_{\mathrm{enc},i}}
\newcommand{\featureextractor}{f_{\mathrm{feat}}}
\newcommand{\featureextractori}{f_{\mathrm{feat},i}}
\newcommand{\experimentheight}{0.67in}
\newcommand{\qheading}[1]{\textbf{#1}}

\begin{abstract}
	The estimation of 3D face shape from a single image must be robust to variations in lighting, head pose, expression, facial hair, makeup, and occlusions.
Robustness requires a large training set of in-the-wild images, which
by construction, lack ground truth 3D shape.
To train a network {\em without any 2D-to-3D supervision}, we present {\em RingNet}, which learns to compute 3D face shape from a single image.
Our key observation is that an individual's face shape is constant across images, regardless of expression, pose, lighting, etc.
RingNet leverages multiple images of a person and automatically detected 2D face features.
It uses a novel loss that encourages the face shape to be similar when the identity is the same and different for different people.
We achieve invariance to expression by representing the face using the FLAME model.
Once trained, our method takes a single image and outputs the parameters of FLAME, which can be readily animated.
Additionally we create a new database of faces {\em ``not quite in-the-wild'' (NoW)}
with 3D head scans and high-resolution images of the subjects in a wide variety of conditions.
We evaluate publicly available methods and find that RingNet is  {\em more accurate than methods that use 3D supervision.}
The dataset, model, and results are available for research purposes at \url{http://ringnet.is.tuebingen.mpg.de}.
   
\end{abstract}

%%%%%%%%% BODY TEXT
\section{Introduction}
Our goal is to estimate 3D head and face shape from a single image of a person.
In contrast to previous methods, we are interested in more than just a tightly cropped region around the face. Instead, we estimate the full 3D face, head and neck.
Such a representation is necessary for applications in VR/AR, virtual glasses try-on, animation, biometrics, etc.
Furthermore, we seek a representation that captures the 3D facial expression, factors face shape from expression, and can be reposed and animated.
While there have been numerous methods proposed in the computer vision literature to address the problem of facial shape estimation \cite{Zollhofer1}, no previous methods address all of our goals.

Specifically, we train a neural network that regresses from image pixels directly to the parameters of a 3D face model.
Here we use FLAME \cite{Bolkart1} because it is more accurate than other models, captures a wide range of shapes, models the whole head and neck, can be easily animated, and is freely available.
Training a network to solve this problem, however, is challenging because there is little paired data of 3D heads/faces together with natural images of people.
For robustness to imaging conditions, pose, facial hair, camera noise, lighting, etc., we wish to train from a large corpus of in-the-wild images.
Such images, by definition, lack controlled ground truth 3D data.

This is a generic problem in computer vision -- finding 2D training data is easy but learning to regress 3D from 2D is hard when paired 3D training data  is very limited and difficult to acquire.
Without ground truth 3D, there are several options but each has problems.  
Synthetic training data typically does not capture real-world complexity.
One can fit a 3D model to 2D image features but this mapping is ambiguous and, consequently, inaccurate.
Because of the ambiguity, training a neural network using only a loss between observed 2D, and projected 3D, features does not lead to good results (cf. \cite{kanazawaHMR18}).

To address the lack of training data, we propose a new method that learns the mapping from pixels to 3D shape {\em without any supervised 2D-to-3D training data.}
To do so, we learn the mapping using only 2D facial features, automatically extracted with OpenPose~\cite{simon2017hand}.
To make this possible, our key observation is that multiple images of the same person provide strong constraints on 3D face shape because the shape remains constant although other things may change such as pose, lighting, and expression.
FLAME factors pose and shape, allowing our model to learn what is constant (shape) and factor out what changes (pose and expression).

While it is a fact that face shape is constant for an individual across images, we need to define a training approach that lets a neural network exploit this shape constancy.
To that end, we introduce {\em RingNet}.
RingNet takes multiple images of a person and enforces that the shape should be similar between all pairs of images, while minimizing the 2D error between observed features and projected 3D features.
While this encourages the network to encode the shapes similarly, we find this is not sufficient.
We also add to the ``ring'' a face belonging to a different random person and enforce that the distance in the latent space between all other images in the ring is larger than the distance between the same person.
Similar ideas have been used in manifold learning (e.g.~triplet loss) \cite{Weinberger} and face recognition \cite{Schroff1}, but, to our knowledge, our approach has not previously been used to learn a mapping from 2D to 3D geometry.
We find that going beyond a triplet to a larger ring, is critical in learning accurate geometry.

While we train with multiple images of a person, note that, at run time, we only need a single image.
With this formulation, we are able to train a network to regress the parameters of FLAME directly from image pixels.
Because we train this with ``in the wild'' images, the network is robust across a wide range of conditions as illustrated in Fig.~\ref{fig:teaser}.
The approach is more general, however, and could be applied to other 2D-to-3D learning problems.

\begin{figure}
%\centerline{\vspace{1.5in}}
	\centerline{
		\includegraphics[height=0.09\textwidth]{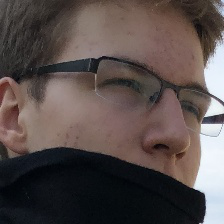} 
		\includegraphics[height=0.09\textwidth]{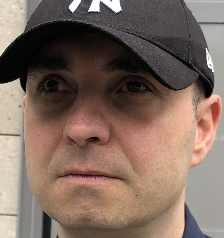}	
		\includegraphics[height=0.09\textwidth]{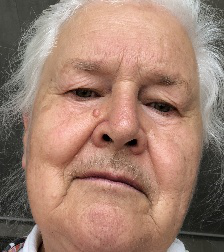}
		\includegraphics[height=0.09\textwidth]{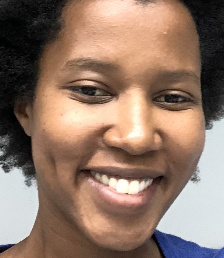} 		
	} 
	\centerline{
		\includegraphics[height=0.09\textwidth]{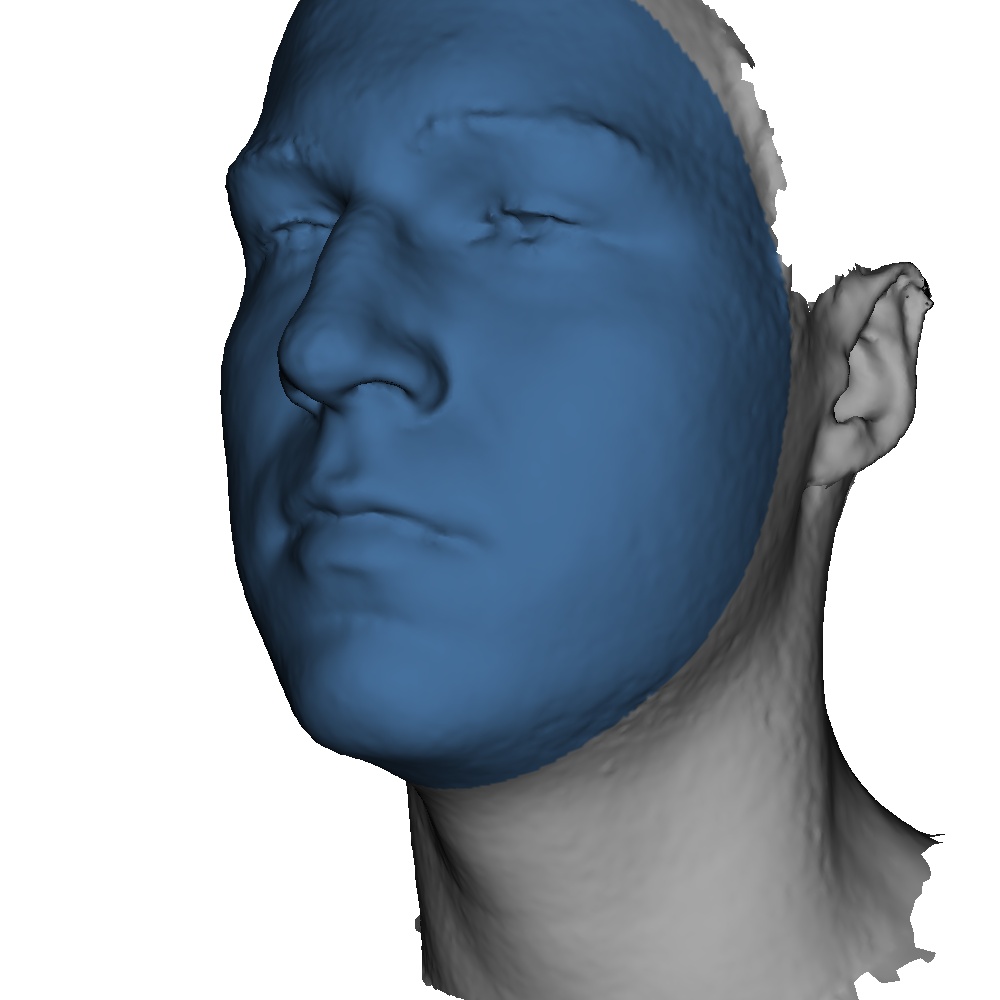} 
		\includegraphics[height=0.09\textwidth]{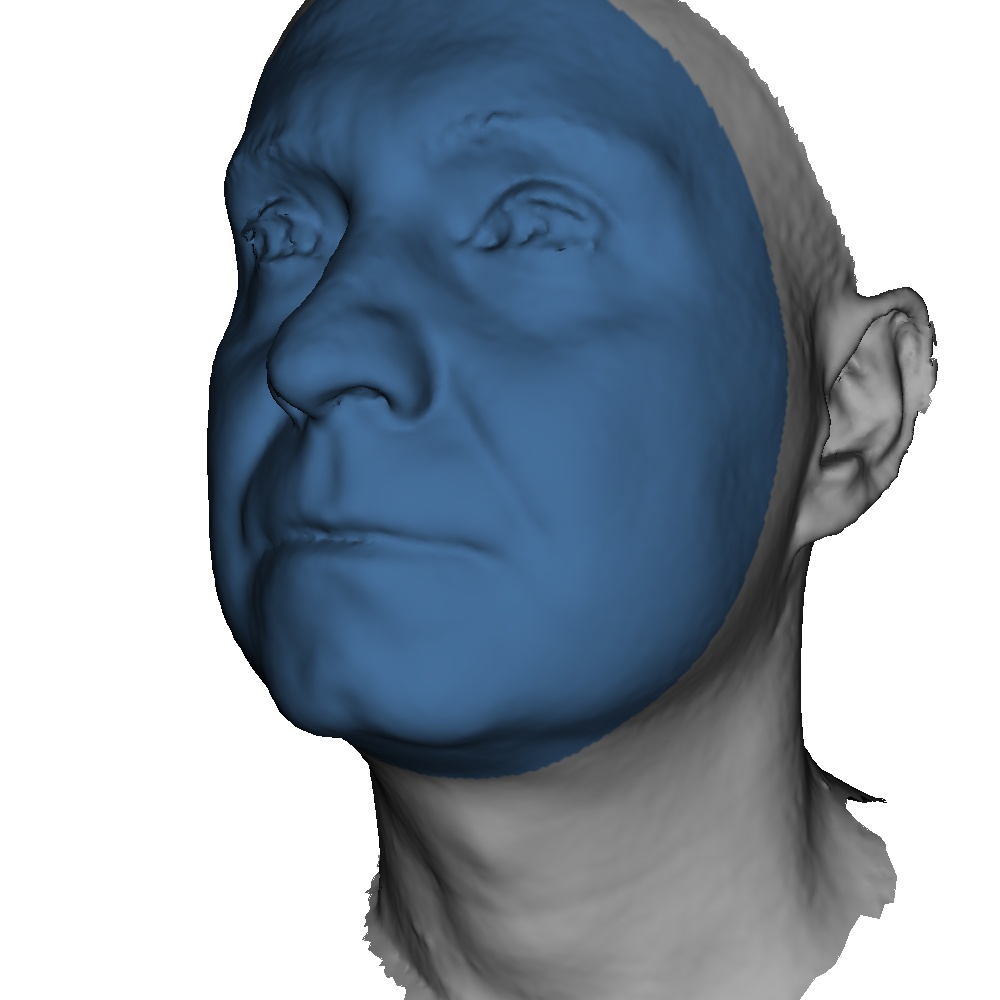}
		\includegraphics[height=0.09\textwidth]{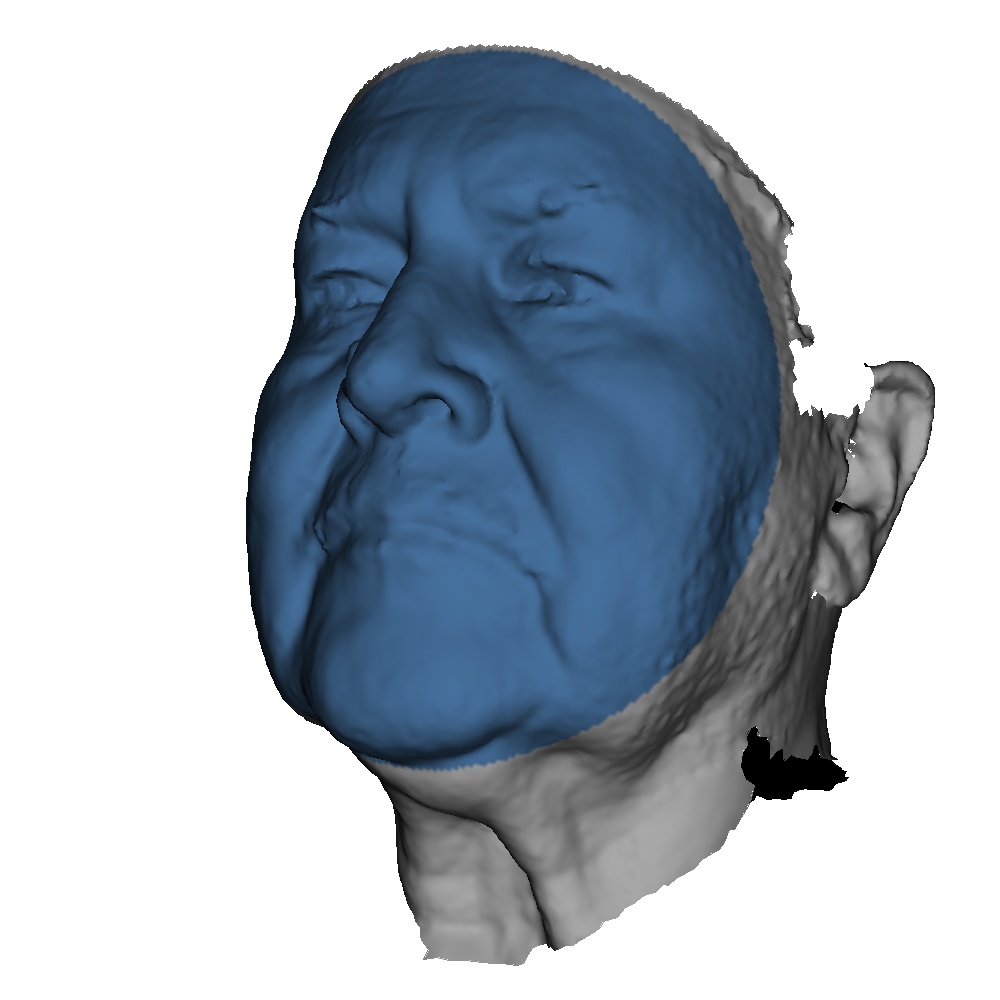}  	
		\includegraphics[height=0.09\textwidth]{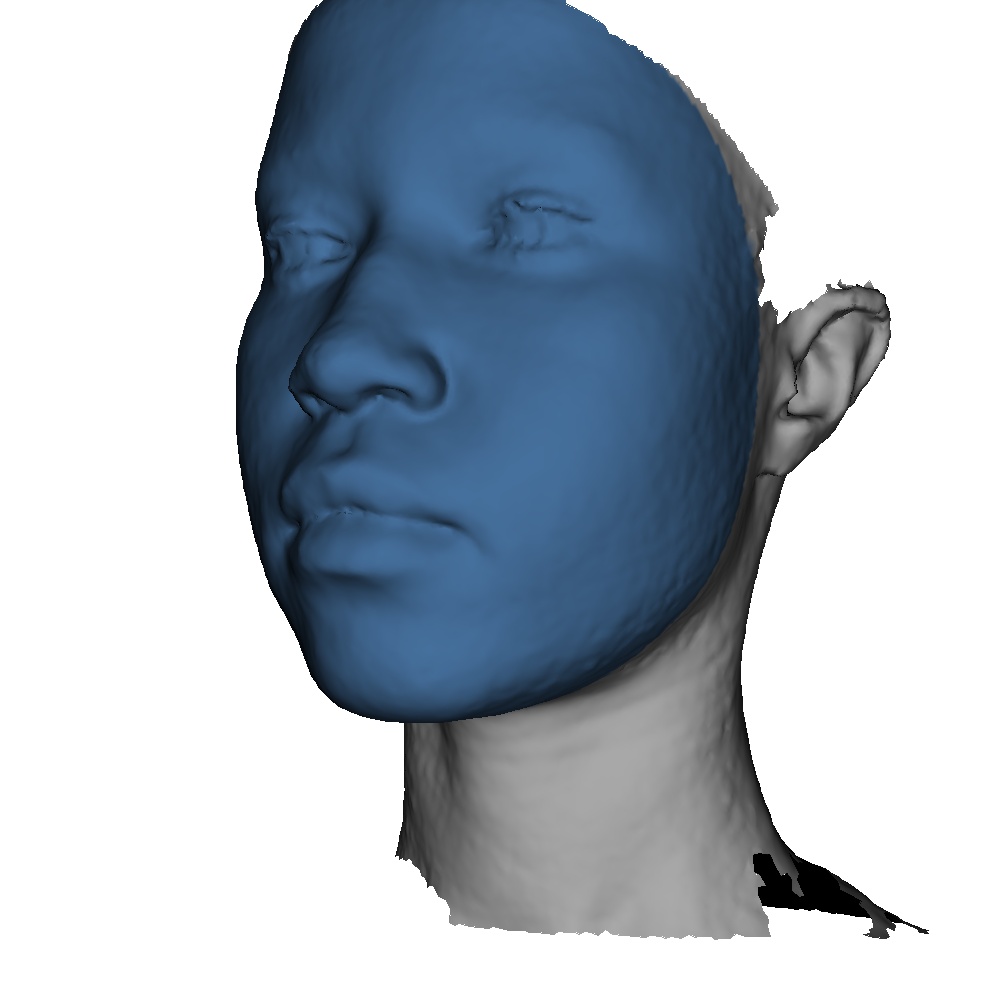} 			
	}
\caption{The {\bf \dataset\ dataset} includes a variety of images take in different conditions (top) and high-resolution 3D head scans (bottom). The dark blue region is the part we considered for face challenge.
}
\label{fig:famosteaser}
\end{figure}

Evaluating the accuracy of 3D face estimation methods remains a challenge and, despite many methods that have been published, there are no rigorous comparisons of 3D accuracy across a wide range of imaging conditions, poses, lighting and occlusion.
To address this, we collected a new dataset called {\em NoW (Not quite in-the-Wild)},
with high-resolution ground truth scans and high-quality images of $100$ subjects taken in a range of conditions (Fig.~\ref{fig:famosteaser}).
\dataset\ is more complex than previous datasets and we use it to evaluate all recent methods with publicly available implementations.
Specifically we compare with \cite{Tran1}, \cite{Tran2} and \cite{Feng}, which are trained with 3D supervision.
Despite not having any 2D-to-3D supervision our RingNet method recovers more accurate 3D face shape.
We also evaluate the method qualitatively on challenging in-the-wild face images.

In summary, the main contributions of our paper are:
(1) Full face, head with neck reconstruction from a single face image.
(2) RingNet -- an end-to-end trainable network that enforces shape consistency across face images of the subject with varying viewing angle, light conditions, resolution and occlusion.
(3) A novel shape consistency loss  for learning 3D geometry from 2D input. 
(4) \dataset\ -- a benchmark dataset for qualitative and quantitative evaluation of 3D face reconstruction methods.
(5) Finally, we  make the model, training code, and new dataset freely available for research purposes to encourage quantitative comparison \cite{Withhold}.

\section{Related work}
There are several approaches to the problem of 3D face estimation from images. One approach estimates depth maps, normals, etc.; that is, these methods produce a representation of object shape tied to pixels but specialized for faces.  The other approach estimates a 3D shape model that can be animated.  We focus on methods in the latter category.
In a recent review paper, Zollh{\"o}fer et al.~\cite{Zollhofer1} describe the state of the art in monocular face reconstruction and provide a forward-looking set of challenges for the field.
Note, that the boundary between supervised, weakly supervised, and unsupervised methods is a blurry one. 
Most methods use some form of 3D shape model, which is learned from scans in advance; we do not call this supervision here.
Here the term supervised implies that paired 2D-to-3D data is used; this might be from real data or synthetic data.
If a 3D model is first optimized to fit 2D image features, then we say this uses 2D-to-3D supervision.
If 2D image features are used but there is no 3D data in training the network, then this is weakly supervised in general and unsupervised relative to the 2D-to-3D task.

\qheading{Quantitative evaluation:} 
Quantitative comparison between methods has been limited by a lack of common datasets with complex images and high-quality ground truth.
Recently, Feng et al.~\cite{Feng2} organized a single image to 3D face reconstruction challenge where they provided the ground truth scans for subjects.
Our \dataset benchmark is complementary to this method as its focus is on extreme viewing angles, facial expressions, and partial occlusions.

\qheading{Optimization:}
Most existing methods require tightly cropped input images and/or reconstruct only a tightly cropped region of the face for which existing shape priors are appropriate.
Most current shape models are descendants of the original Blanz and Vetter 3D morphable model (3DMM) \cite{Blanz1}.
While there are many variations and improvements to this model such as \cite{Gerig}, we use FLAME \cite{Bolkart1} here because both the shape space and expression space are trained from more scans than other methods.
Only FLAME includes the neck region in the shape space and models the pose-dependent deformations of the neck with head rotation. 
Tightly cropped face regions make the estimation of head rotation ambiguous.
Until very recently, this has been the dominant paradigm~\cite{Bolkart2, Suwajanakorn1, Garrido2}.
For example, Kemelmacher-Shlizerman and Seitz \cite{Kemelmacher1} use multi-image shading to reconstruct from collection of images allowing changes in viewpoint and shape.
Thies et al.~\cite{Thies1} achieve accurate results on monocular video sequences.
While these approaches can achieve good results with high-realism, they are computationally expensive.

\qheading{Learning with 3D supervision:}
Deep learning methods are quickly replacing the optimization-based approaches~\cite{Tran2, Zhu1, Kim1, jackson1}.
For example, Sela et al.~\cite{Sela1} use a synthetic dataset to generate an image-to-depth mapping and a pixel-to-vertex mapping, which are combined to generate the face mesh. 
Tran et al.~\cite{Tran1} directly regress the 3DMM parameters of a face model with a dense network. 
Their key idea is to use multiple images of the same subject and fit a 3DMM to each image using 2D landmarks. 
They then take a weighted average of the fitted meshes to use it as the ground truth to train their network.
Feng et al.~\cite{Feng} regress from image to a UV position map that records the position information of the 3D face and provides dense correspondence to the semantic meaning of each point on UV space.
All the aforementioned methods use some form of 3D supervision like synthetic rendering, optimization-based fitting of a 3DMM,  or a 3DMM to generate UV maps or volumetric representation.
None of the fitting-based methods produce true ground truth for real world face images, while synthetically generated faces may not generalize well to the real world \cite{Tewari2}.
Methods that rely on fitting a 3DMM to images using 2D-3D correspondences to create a pseudo ground truth are always limited by the expressiveness of the 3DMM and the accuracy of the fitting process.

\qheading{Learning with weak 3D supervision:} 
Sengupta et al.~\cite{Sengupta} learn to mimic a Lambertian rendering process by using a mixture of synthetically rendered images and real images.
They work with tightly cropped faces and do not produce a model that can be animated. 
Genova  et al.~\cite{Genova} propose an end-to-end learning approach using a differentiable rendering process.
They also train their encoder using synthetic data and its corresponding 3D parameters.
Tran and Liu \cite{Tran3}  learn a nonlinear 3DMM model by using an analytically differentiable rendering layer and in a weakly supervised fashion with 3D data.

\qheading{Learning with no 3D supervision:}

MoFA \cite{Tewari1} estimates the parameters of a 3DMM and is trained end-to-end using a photometric loss and an optional 2D feature loss.
It is effectively a neural network version of the original Blanz and Vetter model in that it models shape, skin reflectance, and illumination to produce a realistic image that is matched to the input.
The advantage of this is that the approach is significantly faster than optimization methods \cite{Tewari2}.
MoFA estimates a tight crop of the face and produces good looking results but has trouble with extreme expressions.
They only perform quantitative evaluation on real images using the FaceWarehouse model as the ``ground truth''; this is not an accurate representation of true 3D face shape.

The methods that learn without any 2D-to-3D supervision all explicitly model the image formation process (like Blanz and Vetter) and formulate a photometric loss and typically also incorporate 2D face feature detections with known correspondence to the 3D model.
The problem with the photometric loss is that the model of image formation is always approximate (e.g. Lambertian).  Ideally, one would like a network to learn not just about face shape but about the complexity of real world images and how they relate to shape.
To that end, our RingNet approach uses only the 2D face features and no photometric term.
Despite (or because of) this, the method is able to learn a mapping from pixels directly to 3D face shape.
This is the least supervised of published methods.

\section{Proposed method}
\label{sec:losses}

\begin{figure}[t]
	\centerline{
		\includegraphics[width=0.8\columnwidth]{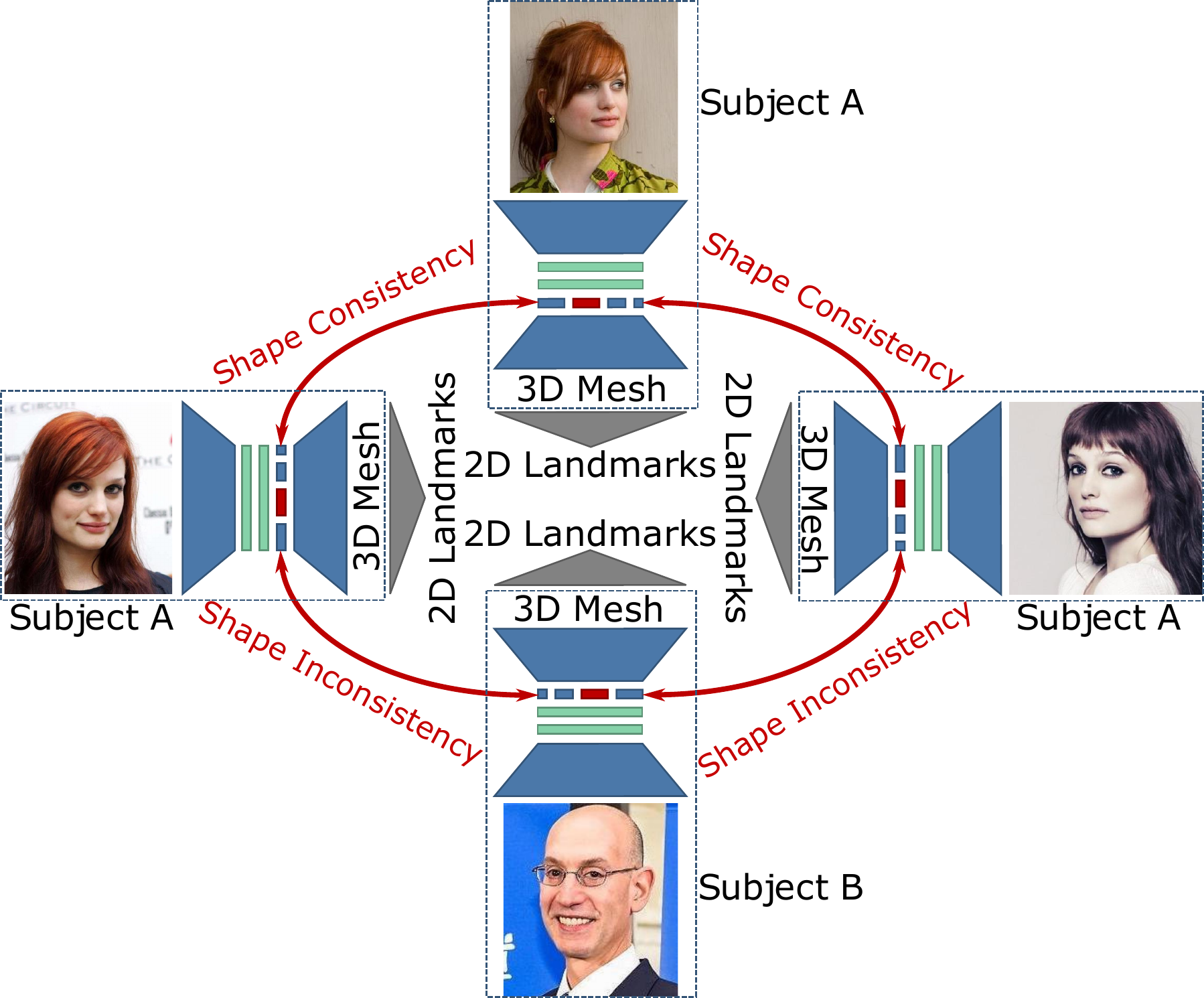} 
	}
	%\centerline{\vspace{1.5in}}
	\caption{RingNet takes multiple images of the same person (Subject A) and an image of a different person (Subject B) during training and enforces shape consistency between the same subjects and shape inconsistency between the different subjects. The computed 3D landmarks from the predicted 3D mesh projected into 2D domain to compute loss with ground-truth 2D landmarks. During inference, RingNet takes a single image as input and predicts the corresponding 3D mesh. Images are taken from~\cite{Cao18}. The figure is a simplified version for illustration purpose.}
	\label{fig:ringnet}
\end{figure}

\begin{figure}[t]
	\centerline{
		\includegraphics[width=0.6\columnwidth]{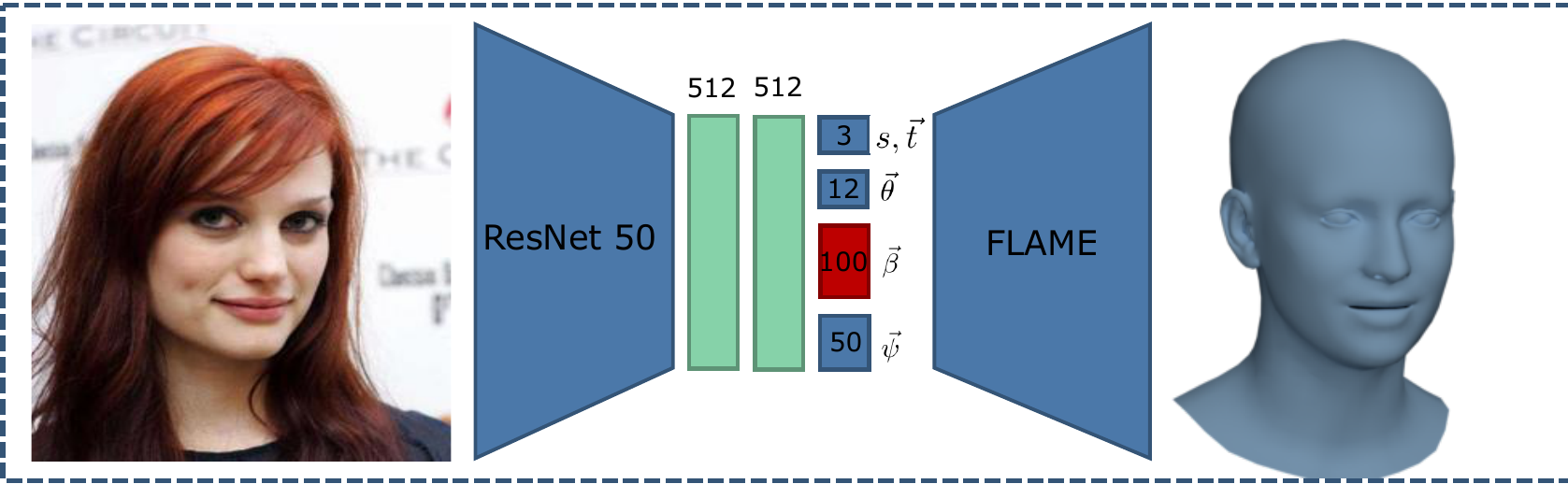} 
	}
	\caption{Ring element that outputs a 3D mesh for an image.}
	\label{fig:ringnet_element}
\end{figure}

The goal of our method is to estimate 3D head and face shape from a single face image $\textit{I}$. 
Given an image, we assume the face is detected, loosely cropped, and approximately centered. 
During training, our method leverages 2D landmarks and identity labels as input.
During inference it uses only image pixels; 2D landmarks and identity labels are not used.

\textbf{Key idea:} The key idea can be summarized as follows: 1) The face shape of a person remains unchanged, even though an image of the face may vary in viewing angle, lighting condition, resolution, occlusion, expression or other factors. 2) Every person has a unique face shape (not considering identical twins).

We leverage this idea by introducing a shape consistency loss, embodied in our ring-structured network. RingNet (Fig.~\ref{fig:ringnet}) is a multiple encoder-decoder based architecture, with weight sharing between the encoders, and shape constraints on the shape variables. Each encoder in the ring is a combination of a feature extractor network and a regressor network. Imposing shape constraints on the shape variables forces the network to disentangle facial shape, expression, head pose, and camera parameters. We use FLAME~\cite{Bolkart1} as a decoder to reconstruct 3D faces from the semantically meaningful embedding, and to obtain a decoupling within the embedding space into semantically meaningful parameters (i.e. shape, expression, and pose parameters). 

We introduce the FLAME decoder, the RingNet architecture, and the losses in more details in the following.

\subsection{FLAME model}

FLAME uses linear transformations to describe identity and expression dependent shape variations, and standard linear blend skinning (LBS) to model neck, jaw, and eyeball rotations around $K = 4$ joints. 
Parametrized by coefficients for shape, $\shapecoeff \in \mathbb{R}^{\shapedim}$, pose $\posecoeff \in \mathbb{R}^{3K+3}$, and expression $\expcoeff \in \mathbb{R}^{\expdim}$, FLAME returns $N = 5023$ vertices. 
FLAME models identity dependent shape variations $B_S(\shapecoeff; \shapespace): \mathbb{R}^{\shapedim} \rightarrow \mathbb{R}^{3N}$, corrective pose blendshapes $B_P(\posecoeff; \posespace): \mathbb{R}^{3K+3} \rightarrow \mathbb{R}^{3N}$, and expression blendshapes $B_E(\expcoeff; \expspace): \mathbb{R}^{\expdim} \rightarrow \mathbb{R}^{3N}$ as linear transformations with learned bases $\shapespace$, $\expspace$, and $\posespace$. 
Given a template $\template \in \mathbb{R}^{3N}$ in the ``zero pose'', identity, pose, and expression blendshapes, are modeled as vertex offsets from $\template$.

Each of the pose vectors $\posecoeff \in \mathbb{R}^{3K+3}$ contains (K+1) rotation vectors in axis-angle representation;  i.e.~one vector per joint plus the global rotation. 
The blend skinning function $W(\template, \joints, \posecoeff, \blendweights)$ then rotates the vertices around the joints $\joints \in \mathbb{R}^{3K}$, linearly smoothed by blendweights $\blendweights \in \mathbb{R}^{K \times N}$. 
More formally, FLAME is given as
\begin{equation}
	M(\shapecoeff, \posecoeff, \expcoeff) = W(T_P(\shapecoeff, \posecoeff, \expcoeff),  \joints(\shapecoeff), \posecoeff, \blendweights),
	\label{eq:flamemodel}
\end{equation}
with
\begin{equation}
	T_P(\shapecoeff, \posecoeff, \expcoeff) = \template + B_S(\shapecoeff; \shapespace) + B_P(\posecoeff; \posespace) + B_E(\expcoeff; \expspace).
\end{equation}
The joints are defined as a function of $\shapecoeff$ since different face shapes require different joint locations. 
We use Equation \ref{eq:flamemodel} for decoding our embedding space to generate a 3D mesh of a complete head and face. 

\subsection{RingNet}

The recent advances in face recognition (e.g.~\cite{Zhang2017}) and facial landmark detection (e.g.~\cite{bulat2017far,simon2017hand}) have led to large image datasets with identity labels and 2D face landmarks. For training, we assume a corpus of 2D face images $I_i$, corresponding identity labels $c_i$, and landmarks $k_i$. %given.
%, denoted as , $k_i$,{tre given as $\{I_i, c_i, k_i\}_{i=1}^{i=N_{\mathrm{train}}}$ given in the following. 

The shape consistency assumption can be formalized by $\shapecoeff_i = \shapecoeff_j, \forall c_i = c_j$ (i.e.~the face shape of one subject should remain the same across multiple images) and $\shapecoeff_i \neq \shapecoeff_j, \forall c_i \neq c_j$ (i.e.~the face shape of different subjects should be distinct). RingNet introduces a ring-shaped architecture that jointly optimizes for shape consistency for an arbitrary number input images in parallel. For details regarding the shape consistency, see Section~\ref{sec:losses}.

RingNet is divided into $\numboferelements$ ring elements $\elemntsofring_{i=1}^{i=\numboferelements}$ as shown in Figure~\ref{fig:ringnet}, where each $\elemntsofring_i$ consists of an encoder and a decoder network (see Figure~\ref{fig:ringnet_element}). The encoders share weights across $\elemntsofring_i$, the decoder weights remain fixed during training. The encoder is a combination of a feature extractor network $\featureextractor$ and regression network $f_{\mathrm{reg}}$. Given an image $I_i$, $\featureextractor$ outputs a high-dimensional vector, which is then encoded by $f_{\mathrm{reg}}$ into a semantically meaningful vector 
(i.e.,~$\encoderoutput(I_i) = f_{\mathrm{reg}}(\featureextractor(I_i))$). This vector can be expressed as a concatenation of the camera, pose, shape and expression parameters, i.e., $\encoderoutput(I_i) = [\mathrm{cam}_i, \posecoeff_i, \shapecoeff_i, \expcoeff_i]$, where $\posecoeff_i, \shapecoeff_i, \expcoeff_i$ are FLAME parameters. 

For simplicity we omit $I$ in the following and use $\encoderoutput(I_i) = {\encoderoutputi}$ and $\featureextractor(I_i) = {\featureextractori}$. The regression network iteratively regresses ${\encoderoutputi}$ in an iterative error feedback loop~\cite{kanazawaHMR18, Carreira1}, instead of directly regressing $\encoderoutputi$ from $\featureextractori$. In each iteration step, progressive shifts from the previous estimate are made to reach the current estimate. Formally the regression network takes the concatenated $[\featureextractori^t, \encoderoutputi^t]$ as input and gives $\delta\encoderoutputi^t$ as output. Then we update the current estimate by,
\begin{equation}
{\encoderoutputi}^{t+1} = {\encoderoutputi}^t + \delta{\encoderoutputi}^t.
\end{equation}
This iterative network performs multiple regression iterations per iteration of the entire RingNet training. The initial estimate is set to $\vec{0}$. The output of the regression network is then fed to the differentiable FLAME decoder network which outputs the 3D head mesh.

The number of ring elements $\numboferelements$ is a hyper-parameter of our network, which determines the number of images processed in parallel with optimized consistency on the $\shapecoeff$. RingNet allows to use any combination of images of the same subject and images of different subjects in parallel. However, without loss of generality, we feed face images of the same identity to $\{\elemntsofring_j\}_{j=1}^{j=\numboferelements-1}$ and different identity to $\elemntsofring_{\numboferelements}$. Hence for each input training batch, each slice consists of $\numboferelements-1$ images of the same person and one image of another person (see Fig.~\ref{fig:ringnet}).

\subsection{Shape consistency loss}

For simplicity let us call two subjects who have same identity label  ``matched pairs'' and two subjects who have different identity labels are ``unmatched pairs''.
A key goal of our work is to make a robust end-to-end trainable network that can produce the same shapes from images of the same subject and different shapes for different subjects.
In other words we want to make our shape generators discriminative. 
%This is achievable by distance metric learning .
We enforce this by requiring matched pairs to have a distance in shape space that is smaller by a margin, $\eta$, than the distance for unmatched pairs.
Distance is computed in the space of face shape parameters, which corresponds to a Euclidean space of vertices in the neutral pose.

In the RingNet structure, $\elemntsofring_j$ and $\elemntsofring_k$ produce $\shapecoeff_j$ and $\shapecoeff_k$, which are matched pairs when $j \neq k$ and $j,k \neq \numboferelements$.
Similarly $\elemntsofring_j$ and $\elemntsofring_{\numboferelements}$ produce $\shapecoeff_j$ and $\shapecoeff_{\numboferelements}$, which are unmatched pairs when $j \neq \numboferelements$.
Our shape constancy term is then
%To enforce robustness we make our RingNet discriminative by enforcing,
\begin{equation}
		\left \| \shapecoeff_j - \shapecoeff_k \right \|_2^2  + \eta \leq \left \| \shapecoeff_j - \shapecoeff_{\numboferelements} \right \|_2^2
\end{equation}
%where $j,k \neq \numboferelements$. 
Thus we minimize the following loss while training RingNet end-to-end, $L_S=$
\begin{equation}
\label{tripletloss}
\sum_{i=1}^{n_b} \sum_{j,k=1}^{\numboferelements-1} \max(0, \left \| \shapecoeff_{ij} - \shapecoeff_{ik} \right \|_2^2 - \left \| \shapecoeff_{ij} - \shapecoeff_{i\numboferelements} \right \|_2^2 + \eta)
\end{equation}
which is normalized to,
\begin{equation}
	L_{\mathrm{SC}}= \frac{1}{n_b \times R} \times L_S
\end{equation}
where $n_b$ is the batch size for each element in the ring.

\subsection{2D feature loss}

Finally we compute the $L_1$ loss between the ground-truth landmarks provided during the training procedure and the predicted landmarks. Note that we do not directly predict 2D landmarks, but 3D meshes with known topology, from which the landmarks are retrieved.

Given the FLAME template mesh, we define for each OpenPose~\cite{simon2017hand} keypoint the corresponding 3D point in the mesh surface.
Note that this is the only place where we provide supervision that connects 2D and 3D. 
This is done only once.
While the mouth, nose, eye, and eyebrow keypoints have a fixed corresponding 3D point (referred to as static 3D landmarks), the position of the contour features changes with head pose (referred to as dynamic 3D landmarks).
Similar to~\cite{Cao:2014,Tewari2}, we model the contour landmarks as dynamically moving with the global head rotation (see Sup. Mat.). 
To automatically compute this dynamic contour, we rotate the FLAME template between -20 and 40 degrees to the left and right, render the mesh with texture, run OpenPose to predict 2D landmarks, and project these 2D points to the 3D surface. 
The resulting trajectories are symmetrically transferred between the left and right side of the face.

During training, RingNet outputs 3D meshes, computes the static and dynamic 3D landmarks for these meshes, and projects these into the image plane using the camera parameters predicted in the encoder output.
Henceforth we compute the following $L_1$ loss between the projected landmarks $k_{p_i}$ and the ground-truth 2D landmarks $k_i$.
\begin{equation}
	L_{\mathrm{proj}} = \left \| w_i \times (k_{p_i} - k_i) \right \|_1
\end{equation} 
where $w_i$ is the confidence score of each ground-truth landmark which is provided by the 2D landmark predictor. We set it to $1$ if the confidence is above $0.41$ and to $0$ otherwise.
The total loss $L_{tot}$, which trains RingNet end-to-end is
\begin{equation}
	\begin{aligned}
		L_{tot} = \lambda_{SC} L_{\mathrm{SC}} + \lambda_{proj} L_{\mathrm{proj}} + \lambda_{\shapecoeff} \left \| \shapecoeff \right \|_2^2 + \lambda_{\expcoeff} \left \| \expcoeff \right \|_2^2 
	\end{aligned}
\end{equation}
where the $\lambda$ are the weights of each loss term and the last two terms regularize the shape and expression coefficients. Since $B_S(\shapecoeff; \shapespace)$ and $B_E(\expcoeff; \expspace)$ are scaled by the squared variance, the L2 norm of $\shapecoeff$ and $\expcoeff$ represent the Mahalanobis distance in the orthogonal shape and expression space.

\subsection{Implementation details}

The feature extractor network uses a pre-trained ResNet-50~\cite{He1} architecture, also optimized during training. The feature extractor network outputs a 2048 dimensional vector. That serves as input to the regression network. The regression network consists of two fully-connected layers of dimension 512 with ReLu activation and dropout, followed by a final linear fully-connected layer with 159-dimensional output. 
To this 159-dimensional output vector we concatenate the camera, pose, shape, and expression parameters. The first three elements represent scale and 2D image translation. The following 6 elements are the global rotation and jaw rotation, each in axis-angle representation. The neck and eyeball rotations of FLAME are not regressed since the facial landmarks do not impose any constraints on the neck. The next 100 elements are the shape parameters, followed by 50 expression parameters of FLAME. The differentiable FLAME layer is kept fixed during training. 
We train RingNet for 10 epochs with a constant learning rate of 1e-4, and use Adam~\cite{kinga2015method} for optimization. The different model parameters are $\numboferelements = 6$, $\lambda_{SC} = 1$, $\lambda_{proj} = 60$, $\lambda_{\shapecoeff} = 1e-4$, $\lambda_{\expcoeff} = 1e-4$, $\eta = 0.5$. The RingNet architecture is implemented in Tensorflow~\cite{abadi2016tensorflow} and will be made publicly available.
We use VGG2 Face database \cite{Cao18} as our training dataset which consists of face images and their corresponding labels.
We run OpenPose  \cite{simon2017hand} on the database and compute 68 landmark points on the face.
OpenPose fails for many cases.
After cleaning for the failed cases we have around 800K images with their corresponding labels and facial landmarks for our training corpus.
We also consider around 3000 extreme pose images with corresponding landmarks provided by \cite{bulat2017far}.
Since for these extreme images we do not have any labels we replicate each image with random crops and scale for matched pair consideration.

\section{Benchmark dataset and evaluation metric}
\label{benchmarkdataset}
This section introduces our \dataset benchmark for the task of 3D face reconstruction from single monocular images. The goal of this benchmark is to introduce a standard evaluation metric to measure the accuracy and robustness of 3D face reconstruction methods under variations in viewing angle, lighting, and common occlusions. 

%%%%%%%%%%%%%%%%%%%%%%%%%%%%%%%%%%%%%%
\textbf{Dataset:} The dataset contains 2054 2D images of 100 subjects, captured with an iPhone X, and a separate 3D head scan for each subject. This head scan serves as ground-truth for the evaluation. The subjects are selected to contain variations in age, BMI, and sex (55 female, 45 male).

We categorize the captured data in four challenges; \textit{neutral} (620 images), \textit{expression} (675 images), \textit{occlusion} (528 images) and \textit{selfie} (231 images). \textit{Neutral}, \textit{expression} and \textit{occlusion} contain neutral, expressive, and partially occluded face images of all subjects in multiple views, ranging from frontal view to profile view. \textit{Expression} contains different acted facial expressions such as happiness, sadness, surprise, disgust, and fear. \textit{Occlusion} contain images with varying occlusions from e.g. glasses, sunglasses, facial hair, hats or hoods. For the \textit{selfie} category, participants are asked to take selfies with the iPhone, without imposing constraints on the performed facial expression. The images are captured indoor and outdoor to provide variations of natural and artificial light. 

The challenge for all categories is to reconstruct a neutral 3D face given a single monocular image. Note that facial expressions are present in several images, which requires methods to disentangle identity and expression to evaluate the quality of the predicted identity.

%%%%%%%%%%%%%%%%%%%%%%%%%%%%%%%%%%%%%%
\textbf{Capture setup:} For each subject we capture a raw head scan in neutral expression with an active stereo system (3dMD LLC, Atlanta). The multi-camera system consists of six gray-scale stereo camera pairs, six color cameras, five speckle pattern projectors, and six white LED panels. The reconstructed 3D geometry contains about 120K vertices for each subject. Each subject wears a hair cap during scanning to avoid occlusions and scanner noise in the face or neck region due to hair.

%%%%%%%%%%%%%%%%%%%%%%%%%%%%%%%%%%%%%%
\textbf{Data processing:} Most existing 3D face reconstruction methods require a localization of the face. To mitigate the influence of this pre-processing step we provide for each image, a bounding box, that covers the face. 
To obtain bounding boxes for all images, we first run a face detector on all images~\cite{Zhang2017}, and then predict keypoints for each detected face~\cite{bulat2017far}. We manually select 2D landmarks for failure cases. We then expand the bounding box of the landmarks to each side by $5\%$ (bottom), $10\%$ (left and right), and $30\%$ to the top to obtain a box covering the entire face including forehead.
%We clean the ground truth scans by two methods. 
For the face challenge, we follow processing protocol similar to~\cite{Feng2}. For each scan, the face center is selected, and the scan is cropped by removing everything outside of a specified radius. The selected radius is subject specific computed as $0.7\times (outer\_eye\_dist + nose\_dist)$ (see Figure \ref{fig:famosteaser}). 

%%%%%%%%%%%%%%%%%%%%%%%%%%%%%%%%%%%%%%
\textbf{Evaluation metric:} Given a single monocular image, the challenge consists of reconstructing a 3D face. Since the predicted meshes occur in different local coordinate systems, the reconstructed 3D mesh is rigidly aligned (rotation, translation, and scaling) to the scan using a set of corresponding landmarks between the prediction and the scan. 
We further perform a rigid alignment based on the scan-to-mesh distance (which is the absolute distance between each scan vertex and the closest point in the mesh surface) between the ground truth scan, and the reconstructed mesh using the landmarks alignment as initialization.
The error for each image is then computed as the scan-to-mesh distance between the ground truth scan, and the reconstructed mesh. Different errors are then reported including cumulative error plots over all distances, median distance, average distance, and standard deviation. 

\textbf{How to participate:} To participate in the challenge, we provide a website~\cite{Withhold} to download the test images, and to upload the reconstruction results and selected landmarks for each registration. The error metrics are then automatically computed and returned. Note that we do not provide the ground truth scans to prevent fine-tuning on the test data.

\section{Experiments}
We evaluate RingNet qualitatively and quantitatively and compare our
results with publicly available methods, namely: PRNet (ECCV 2018 \cite{Feng}),
Extreme3D (CVPR 2018 \cite{Tran2}) and 3DMM-CNN (CVPR 2017 \cite{Tran1}).

\qheading{Quantitative evaluation:}
We compare methods on \cite{Feng2} and our \dataset dataset.

 {\bf Feng et al.~benchmark:} 
Feng et al.~\cite{Feng2} describe a benchmark dataset for evaluating 3D face reconstruction from single images. %2D face image to 3D mesh reconstruction. 
They provide a test dataset, that contains facial images and their 3D
ground truth face scans corresponding to a subset of the Stirling/ESRC 3D face database. 
The test dataset contains  2000 2D neutral face images, including 656
high-quality (HQ) and 1344 low-quality (LQ) images.
The high quality images are taken in controlled scenarios and the low quality images are extracted from video frames.
The data focuses on neutral faces whereas our data has higher variety
in expression, occlusion, and lighting as explained in Section
\ref{benchmarkdataset}.

Recall that the methods we compare with (PRNet, Extreme3D, 3DMM-CNN) use 3D
supervision for training whereas our approach does not. 
PRNet \cite{Feng} requires a very tightly cropped face region to give
good results and performs poorly when given the loosely cropped input image that
comes with the benchmark database (see Sup.~Mat.).
Rather than try to crop the images for PRNet, we run it on the given
images and note when it succeeds:
it outputs meshes for 918 of the low resolution test images
and for 509 of the high-quality images.
To be able to compare with PRNet, we run all the other methods only on
the 1427 images for which PRNet succeeds.

\begin{figure*}[t]
	\centerline{
		\includegraphics[width=0.3\textwidth]{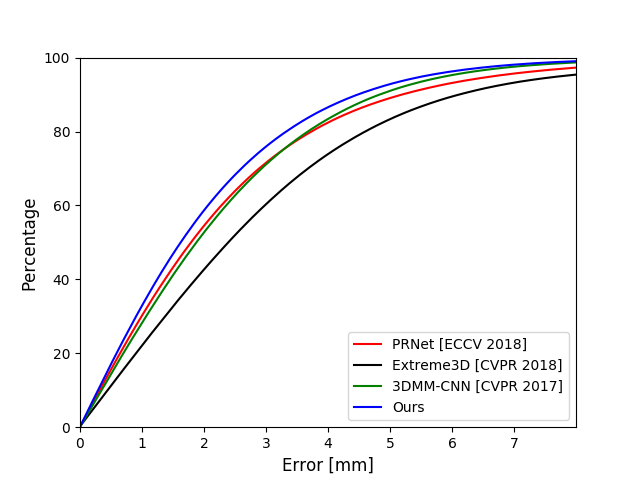}
		\includegraphics[width=0.3\textwidth]{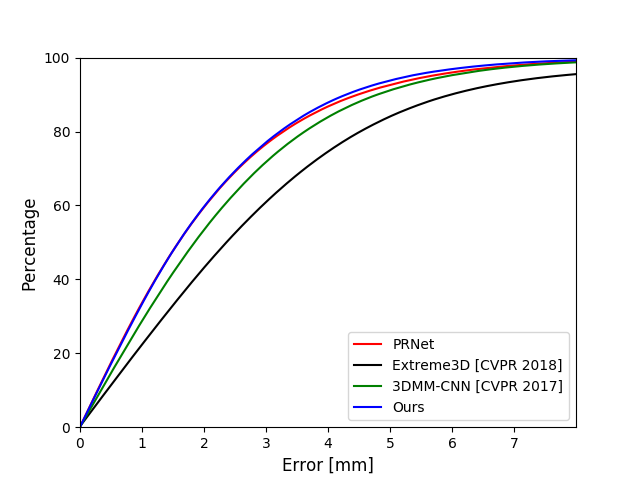}
        \includegraphics[width=0.3\textwidth]{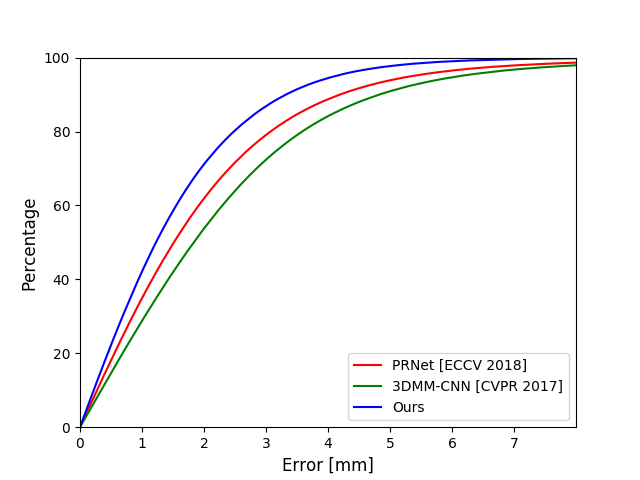}
	}
	\caption{{\bf Cumulative error curves.}  Left to right: LQ
          data of \cite{Feng2}. HQ data of \cite{Feng2}.   \dataset
          dataset face challenge.
	}
	\label{fig:plots}
\end{figure*}
We compute the error using the method in \cite{Feng2}, which computes the distance
from ground truth scan points to the estimated mesh surface.
Figure \ref{fig:plots} (left and middle) show the cumulative error
curve for different approaches for the low-quality and high-quality
images respectively;
RingNet outperforms the other methods.
Table \ref{table:benchmarkdtirling} reports the mean, standard deviation and median errors.

\begin{table}[]
{\footnotesize 
	\begin{tabular}{|l|l|l|l|l|l|l|}
		\hline
		\multirow{2}{*}{\textbf{Method}}                 & \multicolumn{2}{l|}{\textbf{\begin{tabular}[c]{@{}l@{}}Median\\ (mm)\end{tabular}}} & \multicolumn{2}{l|}{\textbf{\begin{tabular}[c]{@{}l@{}}Mean\\ (mm)\end{tabular}}} & \multicolumn{2}{l|}{\textbf{\begin{tabular}[c]{@{}l@{}}Std\\ (mm)\end{tabular}}} \\ \cline{2-7} 
		& \textbf{LQ}                              & \textbf{HQ}                              & \textbf{LQ}                             & \textbf{HQ}                             & \textbf{LQ}                             & \textbf{HQ}                            \\ \hline
		\textbf{PRNet \cite{Feng}}      & 1.79                                     & 1.60                                     & 2.38                                    & 2.06                                    & 2.19                                    & 1.79                                   \\ \hline
		\textbf{Extreme3D \cite{Tran2}} & 2.40                                     & 2.37                                     & 3.49                                    & 3.58                                    & 6.15                                    & 6.75                                   \\ \hline
		\textbf{3DMM-CNN \cite{Tran1}}  & 1.88                                     & 1.85                                     & 2.32                                    & 2.29                                    & 1.89                                    & 1.88                                   \\ \hline
		\textbf{Ours}                                    & \textbf{1.63}                            & \textbf{1.58}                            & \textbf{2.08}                           & \textbf{2.02}                           & \textbf{1.79}                           & \textbf{1.69}                          \\ \hline
	\end{tabular}
}
	\caption{Statistics on Feng et al. \cite{Feng2} benchmark}
	\label{table:benchmarkdtirling}
\end{table}

\textbf{\dataset face challenge:} For this challenge we use cropped scans like \cite{Feng2} to evaluate different methods.
We first perform a rigid alignment of the predicted meshes to the scans
for all the compared methods. 
Then we compute the scan-to-mesh distance \cite{Feng2}  between the
predicted meshes and the scans as above.
Figure \ref{fig:plots} (right) shows the cumulative error curves for
the different methods; again RingNet outperforms the others.
We provide the mean, median and standard division error in Table \ref{table:benchmarkFAMOSFACE}.

\begin{table}[]
	{\footnotesize
		\centerline{
			\begin{tabular}{|l|l|l|l|}
				\hline
				\multicolumn{1}{|c|}{\textbf{Method}} & \multicolumn{1}{c|}{\textbf{\begin{tabular}[c]{@{}c@{}}Median\\ (mm)\end{tabular}}} & \multicolumn{1}{c|}{\textbf{\begin{tabular}[c]{@{}c@{}}Mean\\ (mm)\end{tabular}}} & \multicolumn{1}{c|}{\textbf{\begin{tabular}[c]{@{}c@{}}Std\\ (mm)\end{tabular}}} \\ \hline
				PRNet \cite{Feng} &1.51  &1.99  &1.90  \\ \hline
				3DMM-CNN \cite{Tran1} &1.83  &2.33  &2.05  \\ \hline
				FLAME-neutral \cite{Bolkart1}  &1.24  &1.57  &1.34  \\ \hline
				Ours &\textbf{1.23}  &\textbf{1.55}  &\textbf{1.32}  \\ \hline
			\end{tabular}
		}
	}
	\caption{Statistics for the \dataset dataset face challenge.}
	\label{table:benchmarkFAMOSFACE}
\end{table}

\begin{figure}[tbh]
	\centerline{
		\includegraphics[height=\experimentheight]{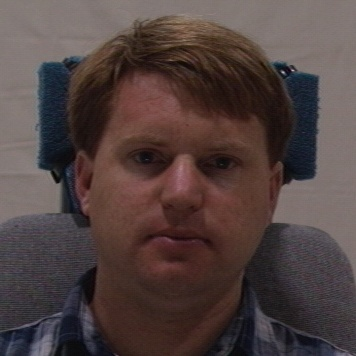} 
		\includegraphics[height=\experimentheight]{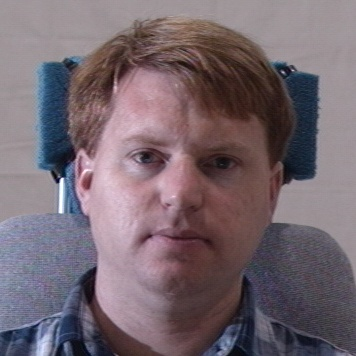} 
		\includegraphics[height=\experimentheight]{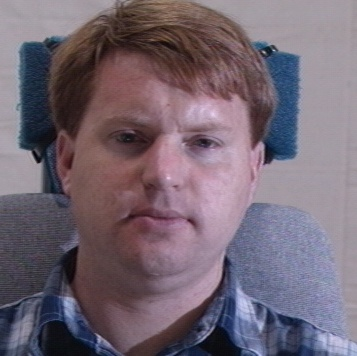} 		
	} 
	\centerline{
		\includegraphics[height=\experimentheight]{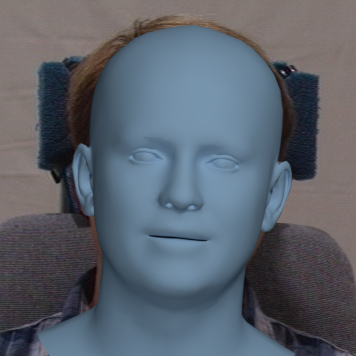} 
		\includegraphics[height=\experimentheight]{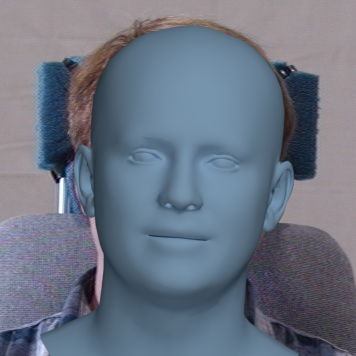} 
		\includegraphics[height=\experimentheight]{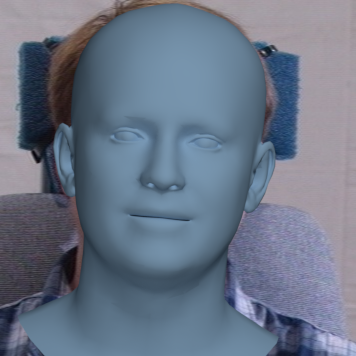} 		
	}
	\caption{Robustness of RingNet to varying
          lighting conditions. Images from the MultiPIE dataset \cite{gross2010multi}.} %We provide more results in the Supplementary material.}
	\label{fig:robustness_towards_lighting}
\end{figure}

\begin{figure}[tbh]
	\centerline{
		\includegraphics[height=\experimentheight]{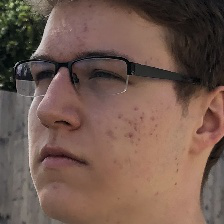} 
		\includegraphics[height=\experimentheight]{IMG_1107_original.png} 
		\includegraphics[height=\experimentheight]{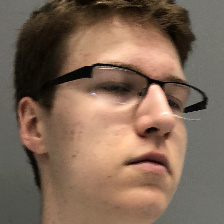} 
	} 
	\centerline{
		\includegraphics[height=\experimentheight]{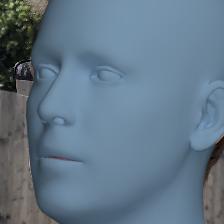} 
		\includegraphics[height=\experimentheight]{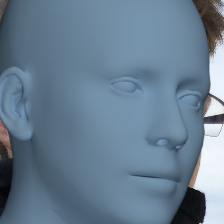} 
		\includegraphics[height=\experimentheight]{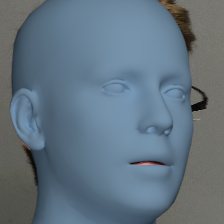} 
	}
	\caption{Robustness of RingNet to occlusions, variations in
		pose, and lighting. Images from the \dataset dataset.}
	\label{fig:robustness_towards_occlusion}
\end{figure}

\qheading{Qualitative results:}
Here we show the qualitative results of estimating a 3D face/head mesh
from a single face image on CelebA \cite{liu2015faceattributes} and MultiPIE dataset \cite{gross2010multi}.
Figure \ref{fig:teaser} shows a few results for RingNet, illustrating
its robustness to expression, gender, head pose, hair, occlusions, etc.
We show robustness of our approach under different conditions like lighting, poses and occlusion in Figures \ref{fig:robustness_towards_lighting} and \ref{fig:robustness_towards_occlusion}. Qualitative comparisons are provided in the Sup.~Mat.~

\qheading{Ablation study:}
Here we provide some motivation for the choice of using a ring architecture in RingNet by comparing different values for R in Table \ref{table:ringelements}. 
We evaluate these on a validation set that contains 2D images and 3D scans of 10 subjects (six subjects from~\cite{Dai_2017_ICCV}, four from \cite{Bolkart1})
For each subject we choose one neutral scan and two to four scanner images, reconstruct the 3D meshes for the images, and measure the scan-to-mesh reconstruction error after rigid alignments. 
%For $R=4$, the errors on the validation set are 2.64mm (mean), 2.22mm (standard deviation), 2.08mm (median), while for $R=3$, the errors are 2.96mm (mean), 2.52mm (standard deviation) and 2.29mm (median).
The error decreases when using a ring structure with more elements over using a single triplet loss only, but it also increases training time.
To make a trade of between time and error, we chose $R=6$ in our experiments.

\begin{table}[]
	{\footnotesize
		\centerline{
			\begin{tabular}{|l|l|l|l|}
				\hline
				\multicolumn{1}{|c|}{\textbf{R}} & \multicolumn{1}{c|}{\textbf{\begin{tabular}[c]{@{}c@{}}Median (mm)\end{tabular}}} & \multicolumn{1}{c|}{\textbf{\begin{tabular}[c]{@{}c@{}}Mean (mm)\end{tabular}}} & \multicolumn{1}{c|}{\textbf{\begin{tabular}[c]{@{}c@{}}Std (mm)\end{tabular}}} \\ \hline
				3 &1.25  &1.68  &1.51  \\ \hline
				4 &1.24  &1.67  &1.50  \\ \hline
				5 &1.20  &1.63  &1.48  \\ \hline
				6 &\textbf{1.19}  &\textbf{1.63}  &\textbf{1.48}  \\ \hline
			\end{tabular}
		}
	}
	\caption{Effect of varying number of ring elements R. We evaluate on a validation set described in the ablation study.}
	\label{table:ringelements}
\end{table}

\section{Conclusion}
We have addressed the challenging problem of learning to estimate a 3D, articulated, and deformable shape from a single 2D image with no paired 3D training data.
We have applied our RingNet model to faces but the formulation is general.
The key idea is to exploit a ring of pairwise losses that encourage the solution to share the same shape for images of the same person and a different shape when they differ.
We exploit the FLAME face model to factor face pose and expression from shape so that RingNet can constrain the shape while letting the other parameters vary.
Our method requires a dataset in which some of the people appear multiple times, as well as 2D facial features, which can be estimated by existing methods.
We provide only the relationship between the standard 2D face features and the vertices of the 3D FLAME model.
Unlike previous methods we do not optimize a 3DMM to 2D features, nor do we use synthetic data.
Competing methods typically exploit a photometric loss using an approximate generative model of facial albedo, reflectance and shading.
RingNet does not need this to learn the relationship between image pixels and 3D shape.
In addition, our formulation captures the full head and its pose.
Finally, we have created a new public dataset with accurate ground truth 3D head shape and high-quality images taken in a wide range of conditions.
Surprisingly, RingNet outperforms methods that use 3D supervision.
This opens many directions for future research, for example extending RingNet with \cite{COMA:ECCV2018}.
Here we focused on a case with no 3D supervision but we could relax this and use supervision when it is available.
We expect that a small amount of supervision would increase accuracy while the large dataset of in-the-wild images provides robustness to illumination, occlusion, etc.
Our 2D feature detector does not include the ears, though these are highly distinctive features.
Adding 2D ear detections would further improve the 3D head pose and shape.
While our model stops with the neck, we plan to extend our model to the full body \cite{SMPL-X:2019}.
It would be interesting to see if RingNet can be extended to reconstruct 3D body pose and shape from images solely using 2D joints.
This could go beyond current methods, like HMR \cite{kanazawaHMR18}, to learn about body shape.
While RingNet learns a mapping to an existing 3D model of the face, we could relax this and also optimize over the low-dimensional shape space, enabling us to learn a more detailed shape model from examples.
For this, incorporating shading cues \cite{Tewari1,Sengupta} would help constrain the problem.

\qheading{Acknowledgement:}
We thank T. Alexiadis in building the NoW dataset, J. Tesch for rendering results, D. Lleshaj for annotations, A. Osman for supplementary video, and S. Tang for useful discussions. 

\qheading{Disclosure:}
Michael J. Black has received research gift funds from Intel, Nvidia, Adobe, Facebook, and Amazon. He is a part-time employee of Amazon and has financial interests in Amazon and Meshcapade GmbH. His research was performed solely at, and funded solely by, MPI.

%\cleardoublepage
%\clearpage
\section*{Appendix}
\begin{figure}[h]
	\centerline{
		\includegraphics[width=0.5\textwidth]{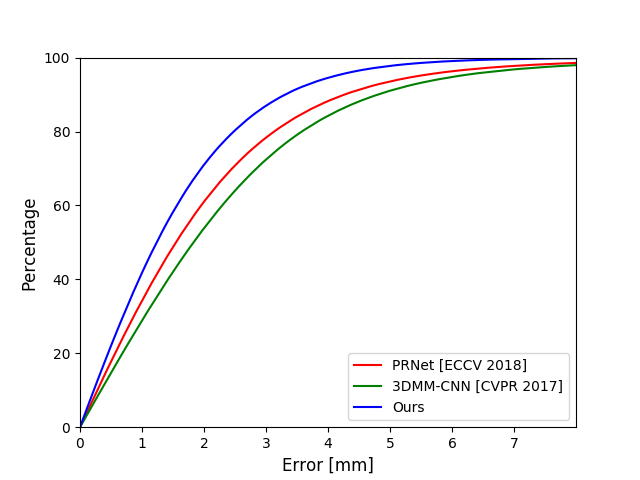} 
 	}
 	\caption{Cumulative error curves for \textit{neutral} challenge.}
 	\label{fig:now_challenge_plots_neutral}
 \end{figure}

\begin{figure}[p]
 	\centerline{
 		\includegraphics[width=0.5\textwidth]{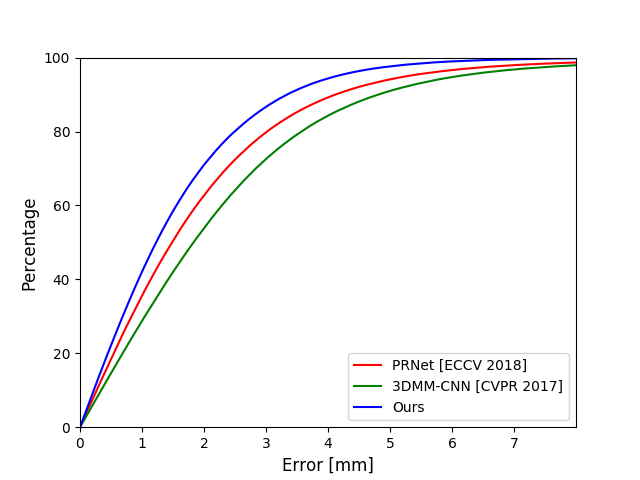} 
 	}
 	\caption{Cumulative error curves for \textit{expression} challenge.}
 	\label{fig:now_challenge_plots_expressions}
	\centerline{
 		\includegraphics[width=0.5\textwidth]{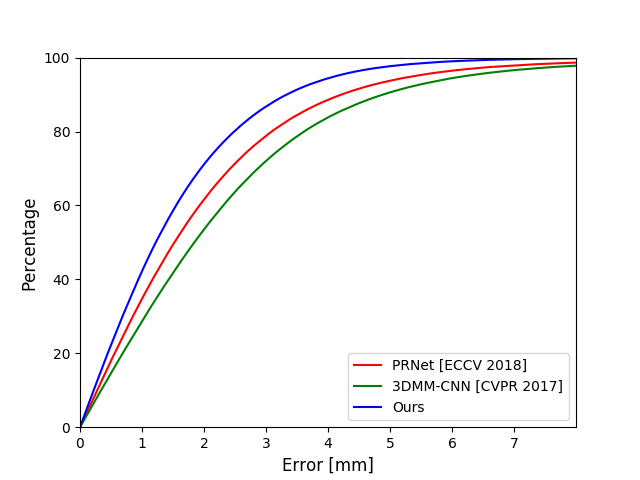}
	}
    \caption{Cumulative error curves for \textit{occlusion} challenge.}
 	\label{fig:now_challenge_plots_occlusions}
	\centerline{
		\includegraphics[width=0.5\textwidth]{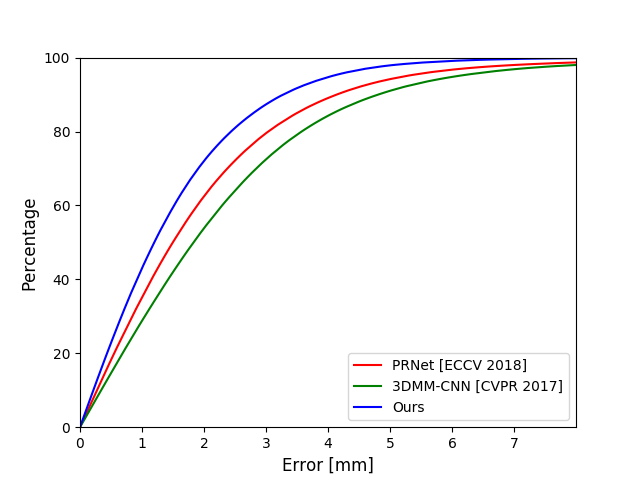}
	}
	\caption{Cumulative error curves for \textit{selfie} challenge.}
	\label{fig:now_challenge_plots_selfie} 	
 \end{figure}

In the following, we show the cumulative error plots for the individual challenges \textit{neutral} (Figure~\ref{fig:now_challenge_plots_neutral}), \textit{expression} (Figure~\ref{fig:now_challenge_plots_expressions}), \textit{occlusion} (Figure~\ref{fig:now_challenge_plots_occlusions}), and \textit{selfie} (Figure~\ref{fig:now_challenge_plots_selfie}) of the NoW dataset. The right of Figure~\ref{fig:plots} shows the cumulative error across all challenges.

%\vfill

\end{document}